\PassOptionsToPackage{numbers,compress}{natbib}

\documentclass{article}

\usepackage[preprint]{neurips_2026}   % nonanonymized arXiv-style preprint
\usepackage[utf8]{inputenc}
\usepackage[T1]{fontenc}
\usepackage{url}
\usepackage{booktabs}
\usepackage{amsfonts}
\usepackage{amsmath,amssymb}
\usepackage{nicefrac}
\usepackage{microtype}
\usepackage[table,xcdraw]{xcolor}  % the [table] option enables \rowcolor used in our tables
\usepackage{xspace}                % required by the abbreviation macros below

\usepackage[export]{adjustbox}
\usepackage{makecell}
\usepackage{tabularx}
\usepackage{array}
\usepackage{multirow}
\usepackage{graphicx}
\usepackage{wrapfig}
\usepackage{subcaption}
\usepackage{amsthm}
\usepackage{pifont}
\usepackage{marvosym}              % provides \Letter glyph for corresponding-author marker

\definecolor{ShadowBlue}{RGB}{242,245,250}
\definecolor{tabmarkyes}{rgb}{0.271,0.769,0.690}
\definecolor{tabmarkno}{rgb}{0.851,0.196,0.196}
\definecolor{tabmarkpartial}{rgb}{0.95,0.77,0.36}
\definecolor{cvprblue}{rgb}{0.21,0.49,0.74}
\definecolor{authorcolor}{RGB}{231, 76, 90}     % rose accent for affiliation indices and envelope
\definecolor{LightGreen}{rgb}{0.45,0.85,0.45}
\definecolor{DarkGrey}{rgb}{0.35,0.35,0.35}
\definecolor{Grey}{rgb}{0.5,0.5,0.5}
\newcommand{\graymidrule}{\arrayrulecolor{gray!50}\midrule\arrayrulecolor{black}}

\newcommand{\cmark}{\textcolor{tabmarkyes}{\ding{51}}}
\newcommand{\xmark}{\textcolor{tabmarkno}{\ding{55}}}
\newcommand{\pmark}{\textcolor{tabmarkpartial}{\ding{51}\rotatebox[origin=c]{-6.2}{\kern-0.7em\ding{55}}}}
\newcolumntype{Y}{>{\raggedright\arraybackslash}X}

\makeatletter
\DeclareRobustCommand\onedot{\futurelet\@let@token\@onedot}
\def\@onedot{\ifx\@let@token.\else.\null\fi\xspace}
 
\def\ie{\emph{i.e}\onedot}

\makeatother

\usepackage[breaklinks,colorlinks,allcolors=cvprblue]{hyperref}

\usepackage{bm}

\def\eqref#1{Eqn.~(\ref{#1})}

\title{ExtraVAR: Stage-Aware RoPE Remapping for Resolution Extrapolation in Visual Autoregressive Models}

\author{
    {\bf \hspace{-10pt}
    Feihong Yan
    $^{{\color{authorcolor}\boldsymbol{1}}}$ \quad
    Shaoyu Liu
    $^{{\color{authorcolor}\boldsymbol{2}}}$ \quad
    Haixuan Wang
    $^{{\color{authorcolor}\boldsymbol{3}}}$ \quad
    Shuai Lu
    $^{{\color{authorcolor}\boldsymbol{1}}}$
    } \\ \vspace{4pt}
    {\bf \hspace{-10pt}
    Linfeng Zhang
    $^{{\color{authorcolor}\boldsymbol{4}}}$ \quad
    Huiqi Li
    $^{\text{\color{authorcolor}\Letter}}$$^{{\color{authorcolor}\boldsymbol{1}}}$ \quad
    Xiangyang Ji
    $^{\text{\color{authorcolor}\Letter}}$$^{{\color{authorcolor}\boldsymbol{5}}}$
    } \\ \vspace{6pt}
    \small
    { \hspace{-10pt}
    $^{\color{authorcolor}\boldsymbol{1}}$ Beijing Institute of Technology \quad
    $^{\color{authorcolor}\boldsymbol{2}}$ Xidian University \quad
    $^{\color{authorcolor}\boldsymbol{3}}$ Northeastern University at Qinhuangdao
    } \\ \vspace{2pt}
    { \hspace{-10pt}
    $^{\color{authorcolor}\boldsymbol{4}}$ Shanghai Jiao Tong University \quad
    $^{\color{authorcolor}\boldsymbol{5}}$ Department of Automation, Tsinghua University
    } \\ \vspace{3pt}
    { \hspace{-10pt}
    $^{\text{\color{authorcolor}\Letter}}$ Corresponding authors:\quad
    \texttt{huiqili@bit.edu.cn} \quad
    \texttt{xyji@tsinghua.edu.cn}
    }
}

\begin{document}
\maketitle

\begin{abstract}
Visual Autoregressive (VAR) models have emerged as a strong alternative to diffusion for image synthesis, yet their fixed training resolution prevents direct generation at higher resolutions. Naively transferring training-free extrapolation methods from LLMs or diffusion models to VAR yields three characteristic failure modes: \emph{global repetition}, \emph{local repetition}, and \emph{detail degradation}. We trace them to a unified \emph{band-stage mismatch}: VAR generates images in a coarse-to-fine, scale-wise process where each stage is driven by a distinct dominant RoPE frequency band, and each failure mode emerges when the dominant band of a particular stage is disrupted. Building on this insight, we propose \textbf{Stage-Aware RoPE Remapping}, a training-free strategy that assigns each frequency band a stage-specific remapping rule, jointly suppressing all three failure modes. We further observe that attention becomes systematically dispersed as the image resolution increases. Existing methods typically depend on predefined attention scaling factors, which are neither adaptive to the target resolution nor capable of faithfully capturing the actual extent of attention dispersion. We therefore propose \textbf{Entropy-Driven Adaptive Attention Calibration}, which quantifies dispersion via a resolution-invariant normalized entropy and yields a closed-form per-head scaling factor that realigns the extrapolated-resolution attention entropy with its training-resolution counterpart. Extensive experiments show that our method consistently outperforms prior resolution-extrapolation methods in both structural coherence and fine-detail fidelity. \emph{Our code is available at \url{https://github.com/feihongyan1/ExtraVAR}.}
\end{abstract}

\section{Introduction}
\label{sec:intro}

Driven by the success of autoregressive (AR) models in natural language processing, the visual-generation community has increasingly embraced the AR paradigm for image synthesis~\citep{van2016pixel,chen2020generative,yu2022scaling,tian2024visual}. Among these efforts, \emph{Visual Autoregressive} (VAR) models~\citep{tian2024visual} have emerged as a competitive alternative to diffusion by reformulating image generation as a coarse-to-fine, next-scale prediction process, achieving strong fidelity and efficient inference. Despite this progress, the capability to synthesize images \emph{beyond the training resolution} remains a fundamental challenge: since the computational cost of self-attention grows at least quadratically with the spatial dimensions, VAR models are trained at a fixed maximum resolution. Given the broad demand for high-resolution imagery in applications such as large-format printing, digital art, and visual content creation, enabling \emph{training-free} resolution extrapolation at inference time has attracted growing interest.

A natural starting point is to borrow training-free extrapolation techniques developed for other paradigms. Positional-encoding remappings such as PI~\citep{chen2023extending}, NTK-aware scaling, and YaRN~\citep{peng2023yarn} are originally designed for LLMs, while training-free extrapolation methods such as RiFlex~\citep{zhao2025riflex} and DyPE~\citep{issachar2025dype} target diffusion-based visual generation. However, applying them directly to VAR results in three failure modes: (i) \emph{global repetition}, where the holistic layout recurs across the image; (ii) \emph{local repetition}, where mid-sized structures, such as individual objects, appear at reduced sizes and in greater numbers than expected; and (iii) \emph{detail degradation}, where fine textures are blurred. To investigate the cause of these failure modes, we begin by analyzing the inherent properties of VAR.

\begin{figure}[t]
    \centering
    \begin{minipage}[t]{0.54\linewidth}
        \centering
        \includegraphics[width=\linewidth]{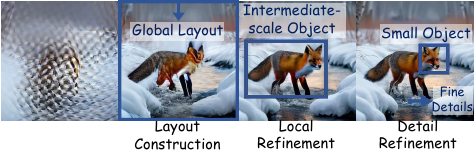}
        \caption{\textbf{Intermediate outputs of VAR at different scale steps.}}
        \label{fig:stage_vis}
    \end{minipage}%
    \hfill
    \begin{minipage}[t]{0.44\linewidth}
        \centering
        \includegraphics[width=\linewidth]{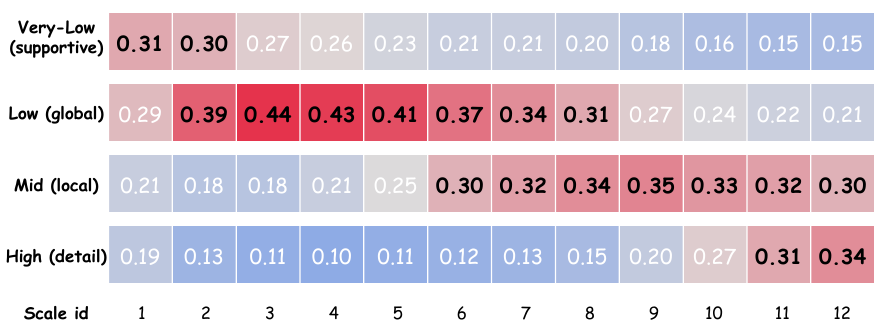}
        \caption{\textbf{Dominant frequency band shifts from low to high in VAR generation.}}
        \label{fig:query_norms}
    \end{minipage}
\end{figure}

\textbf{Image generation in VAR is inherently stage-wise.} Figure~\ref{fig:stage_vis} shows VAR's intermediate outputs across scale steps, revealing a clear coarse-to-fine progression: the generation first forms the overall layout and rough color distribution, then composes mid-sized objects, and finally constructs smaller-sized objects and renders fine-grained details. We partition this process into three stages: the Layout Construction Stage ($k<k_l$), the Local Refinement Stage ($k_l\le k\le k_h$), and the Detail Refinement Stage ($k>k_h$). The above observations suggest that \emph{different generation stages are responsible for fundamentally different tasks and should not be treated uniformly}.

\textbf{Different stages are driven by different RoPE frequency bands.} RoPE encodes position by rotating query and key dimension pairs across a spectrum of frequencies. We partition these pairs by their wavelength $T_j$ relative to the side length $L$ of the training-time 2D token map into four bands: High (the $m$ pairs with smallest $T_j$), Mid ($T_j<L$), Low ($L\le T_j\le 4L$), and Very Low ($T_j>4L$). As shown in Figure~\ref{fig:query_norms}, we quantify each band's contribution to attention by computing the averaged Query $\ell_2$ norm of its rotary pairs at each generation scale step. It can be observed that the dominant band shifts progressively from low through mid to high across the three-stage generation process, indicating that \emph{each stage relies on a distinct frequency band as its primary positional signal}.

\begin{table*}[t]
    \centering
    \footnotesize
    \definecolor{schedulecolor}{rgb}{0.13,0.39,0.70}
    \definecolor{failurecolor}{rgb}{0.78,0.18,0.18}
    \bfseries\itshape
    \caption{\textbf{Stage-wise summary of frequency-band \textbf{Role}, \textcolor{schedulecolor}{\textbf{Schedule}}, and \textcolor{failurecolor}{\textbf{Failure mode}}.}
    $T_j,T'_j$ and $L,L'$ denote wavelengths and 2D token map side lengths at the training and extrapolated resolutions, respectively. ``--'' denotes effects beyond the three analyzed failure modes, whereas \textit{N/A} means the band-stage pair is inactive or non-dominant.}
    \label{tab:freq_roles_by_stage}
    \setlength{\tabcolsep}{5pt}
    \renewcommand{\arraystretch}{1.2}
    \begin{tabularx}{\textwidth}{>{\centering\arraybackslash}m{2.6cm} !{\color{gray!40}\vrule} >{\centering}X !{\color{gray!40}\vrule} >{\centering}X !{\color{gray!40}\vrule} >{\centering}X !{\color{gray!40}\vrule} >{\centering}X}
    \hline
    \toprule
    {\rmfamily\upshape\bfseries Stage}
    & {\rmfamily\upshape\bfseries Very Low (Obs.~1)}
    & {\rmfamily\upshape\bfseries Low (Obs.~2)}
    & {\rmfamily\upshape\bfseries Mid (Obs.~3)}
    & {\rmfamily\upshape\bfseries High (Obs.~4)} \tabularnewline
    \midrule
    {\rmfamily\upshape\bfseries Layout Construction}
        & Global-layout support \\ \textcolor{schedulecolor}{NoPE} \\ \textcolor{failurecolor}{--}
        & Global-layout establishment \\ \textcolor{schedulecolor}{$T_j' = T_j\,L'/L$} \\ \textcolor{failurecolor}{Global repetition}
        & Mid-sized object composition \\ \textcolor{schedulecolor}{$T_j' = T_j\,L'/L$} \\ \textcolor{failurecolor}{Local repetition}
        & \textit{N/A} \\ \textcolor{schedulecolor}{$T_j' = T_j\,L'/L$} \\ \textcolor{failurecolor}{--} \tabularnewline
    \noalign{\color{gray!40}\hrule height 0.4pt}
    {\rmfamily\upshape\bfseries Local Refinement}
        & Mid-sized object support \\ \textcolor{schedulecolor}{NoPE} \\ \textcolor{failurecolor}{--}
        & Global-layout preservation \\ \textcolor{schedulecolor}{$T_j' = T_j\,L'/L$} \\ \textcolor{failurecolor}{--}
        & Mid-sized object composition \\ \textcolor{schedulecolor}{Monotonically decreasing} \\ \textcolor{failurecolor}{Local repetition}
        & \textit{N/A} \\ \textcolor{schedulecolor}{Monotonically decreasing} \\ \textcolor{failurecolor}{--} \tabularnewline
    \noalign{\color{gray!40}\hrule height 0.4pt}
    {\rmfamily\upshape\bfseries Detail Refinement}
        & Fine-detail support \\ \textcolor{schedulecolor}{NoPE} \\ \textcolor{failurecolor}{--}
        & Global-layout preservation \\ \textcolor{schedulecolor}{$T_j' = T_j\,L'/L$} \\ \textcolor{failurecolor}{--}
        & Smaller-sized object composition \\ \textcolor{schedulecolor}{$T_j' \in [T_j,\,T_j\,L'/L]$} \\ \textcolor{failurecolor}{--}
        & Fine-detail rendering \\ \textcolor{schedulecolor}{$T_j' = T_j$} \\ \textcolor{failurecolor}{Detail degradation} \tabularnewline
    \bottomrule
    \hline
    \end{tabularx}
\end{table*}

\textbf{Resolution-extrapolation failures stem from a band-stage mismatch.} To identify the cause of each failure mode, we perform controlled dimension-wise RoPE interventions on the four frequency bands across the three stages. We find that each failure arises from a specific band-stage disruption: perturbing low frequencies in the Layout Construction Stage causes \emph{global repetition}; distorting mid frequencies in the first two stages causes \emph{local repetition}; and suppressing high frequencies in the Detail Refinement Stage causes \emph{detail degradation}. Table~\ref{tab:freq_roles_by_stage} further summarizes the role of each frequency band across the three stages, from which we observe that the role of each band tends to vary from stage to stage. This stage-dependent behavior indicates that \emph{a single static RoPE remapping cannot satisfy all stages at once; instead, the remapping should be adjusted stage by stage to match the changing role of each frequency band}.

Based on the above observations, we propose \textbf{Stage-Aware RoPE Remapping}, a training-free strategy whose remapping scheme evolves to match the dominant frequency band of each stage. Specifically, it enforces global-layout coherence at the \emph{Layout Construction Stage}, reinforces mid-scale structural composition at the \emph{Local Refinement Stage}, and strengthens fine-grained positional discrimination at the \emph{Detail Refinement Stage}, thereby jointly suppressing all three failure modes.

In addition, attention disperses at higher resolution. Prior methods~\citep{peng2023yarn, zhao2025ultraimage} rescale attention with hand-tuned coefficients and quantify dispersion by raw Shannon entropy, which grows with the token count $N$ and is therefore incomparable across resolutions. We instead propose an \textbf{Entropy-Driven Adaptive Attention Calibration}, which adopts a resolution-invariant normalized entropy and derives a closed-form per-head scaling factor that matches its training-resolution reference.

Our contributions are summarized as follows:
\emph{(i)} We identify three characteristic failure modes of VAR resolution extrapolation, namely \emph{global repetition}, \emph{local repetition}, and \emph{detail degradation}, and trace each one to a distinct band-stage disruption.
\emph{(ii)} We propose a \emph{Stage-Aware RoPE Remapping}, which assigns each frequency band a stage-specific remapping rule in order to align the RoPE with the positional requirements of different stages at extended resolution.
\emph{(iii)} We propose an \emph{Entropy-Driven Adaptive Attention Calibration}, which calibrates attention via a resolution-invariant normalized entropy and a closed-form per-head scaling factor.
\emph{(iv)} Extensive experiments on standard benchmarks show that our method consistently outperforms prior resolution-extrapolation methods in both structural coherence and fine-detail fidelity.

%%%%%%%%%%%%%%%%%%%%%%%%%%%%%%%%%%%%%%%%%%%%%%%%%%%%%%%%%%%%%%%%%%%

\begin{table*}[t]
\centering
\small
\caption{\textbf{Comparison of representative training-free resolution-extrapolation methods.} \cmark, \pmark, and \xmark\ indicate that the criterion is explicitly, partially, or not addressed.}
\label{tab:method_comparison}
\setlength{\tabcolsep}{3.5pt}
\renewcommand{\arraystretch}{1.18}
\resizebox{\textwidth}{!}{
\begin{tabular}{l c ccc cc c}
\toprule
\multirow{2}{*}{\textbf{Method}}
& \multirow{2}{*}{\makecell{\textbf{Generation}\\\textbf{Paradigm}}}
& \multicolumn{3}{c}{\textbf{Failure Modes \& Analysis (Sec.~\ref{sec:rope_freq_analysis})}}
& \multicolumn{2}{c}{\textbf{Stage-Aware RoPE (Sec.~\ref{sec:dynamic_rope})}}
& \multirow{2}{*}{\makecell{\textbf{Resolution-invariant}\\\textbf{Entropy (Sec.~\ref{sec:eaac})}}} \\
\cmidrule(lr){3-5}\cmidrule(lr){6-7}
&
&
\makecell{\textbf{Global}\\\textbf{Repetition}}
&
\makecell{\textbf{Local}\\\textbf{Repetition}}
&
\makecell{\textbf{Detail}\\\textbf{Degradation}}
&
\makecell{\textbf{Stage-aware}}
&
\makecell{\textbf{Frequency-aware}}
\\
\midrule
PI~\citep{chen2023extending}
& \textit{LLM}
& \xmark & \xmark & \xmark
& \xmark
& \xmark
& \xmark
\\
YaRN~\citep{peng2023yarn}
& \textit{LLM}
& \xmark & \xmark & \xmark
& \xmark
& \cmark
& \xmark
\\
\makecell[l]{Entropy-aware ABF~\citep{zhang2024extending}}
& \textit{LLM}
& \xmark & \xmark & \xmark
& \xmark
& \pmark
& \pmark
\\
Ms-PoE~\citep{zhang2024found}
& \textit{LLM}
& \xmark & \xmark & \xmark
& \xmark
& \xmark
& \xmark
\\
I-Max~\citep{Du2024IMaxMT}
& \textit{flow}
& \xmark & \xmark & \pmark
& \pmark
& \xmark
& \xmark
\\
HiFlow~\citep{bu2025hiflow}
& \textit{flow}
& \xmark & \xmark & \pmark
& \xmark
& \xmark
& \xmark
\\
Attn-SF~\citep{jin2023training}
& \textit{diffusion}
& \pmark & \xmark & \xmark
& \xmark
& \xmark
& \xmark
\\
RiFlex~\citep{zhao2025riflex}
& \textit{video diffusion}
& \cmark & \xmark & \cmark
& \xmark
& \pmark
& \xmark
\\
InfoScale~\citep{zhang2025infoscale}
& \textit{diffusion}
& \cmark & \pmark & \cmark
& \pmark
& \pmark
& \xmark
\\
DyPE~\citep{issachar2025dype}
& \textit{diffusion}
& \xmark & \xmark & \pmark
& \pmark
& \xmark
& \xmark
\\
\specialrule{0.6pt}{0pt}{2pt}
\rowcolor{ShadowBlue}
\textbf{ExtraVAR (ours)}
& \textit{\textbf{VAR}}
& \cmark & \cmark & \cmark
& \cmark
& \cmark
& \cmark
\\
\bottomrule
\end{tabular}
}
\vspace{4pt}
\end{table*}

\section{Related Work}

\subsection{Inference-Time Attention Recalibration}

Training-free attention recalibration has been widely explored to improve out-of-distribution generation. For long-context language modeling, YaRN~\citep{peng2023yarn} introduces temperature-based attention scaling, which is further refined by position- and layer-dependent rescaling~\citep{zhang2024extending}; theoretical analyses suggest a logarithmic scaling rule with respect to context length~\citep{chen2025critical} and identify attention uncertainty as a central cause of long-context retrieval failure~\citep{zhong2025understanding}. In the visual domain, Attn-SF~\citep{jin2023training} attributes resolution failures to token-count-induced entropy shifts, AlignVid~\citep{liu2025alignvid} applies selective Q/K scaling to mitigate semantic negligence in video generation, Scalable-Softmax~\citep{nakanishi2025scalable} redesigns the normalization kernel to prevent attention fading, InfoScale~\citep{zhang2025infoscale} couples adaptive attention aggregation with frequency compensation, and ERG~\citep{ifriqi2025entropy} replaces the unconditional branch of classifier-free guidance with a perturbed attention distribution.

\subsection{Higher-Resolution Image Generation}

Generating images at resolutions substantially beyond the training resolution remains difficult, often producing repeated patterns, distorted geometry, or weakened coherence. Training-based methods finetune models at higher resolutions or rely on cascaded and size-conditioned pipelines~\citep{hoogeboom2023simple,teng2023relay,zheng2024any,guo2024make,ren2024ultrapixel,yu2025ultra,zhang2025diffusion,xie2024sana,wang2025native,zhang2025ledit}, but typically demand substantial compute and high-quality high-resolution data. Training-free alternatives keep the weights frozen and adapt at inference time: tile-based decomposition preserves local coherence~\citep{BarTal2023MultiDiffusionFD,lee2023syncdiffusion}; convolution or attention restructuring enlarges the effective receptive field~\citep{he2023scalecrafter,haji2024elasticdiffusion,zhang2024hidiffusion,kim2025diffusehigh}; frequency-aware corrections suppress structural artifacts~\citep{huang2024fouriscale,Du2024IMaxMT,zhang2024frecas}; and finer-grained attention or feature recalibration further improves robustness~\citep{qiu2025freescale,zhang2025infoscale,tan2025freepca,koh2025scalediff}. Table~\ref{tab:method_comparison} contrasts representative training-free methods with our method.

\section{Method}

\subsection{Frequency Component Analysis in RoPE}
\label{sec:rope_freq_analysis}

\begin{figure}[t]
    \centering
    \begin{minipage}[t]{0.48\linewidth}
        \centering
        \includegraphics[width=\linewidth]{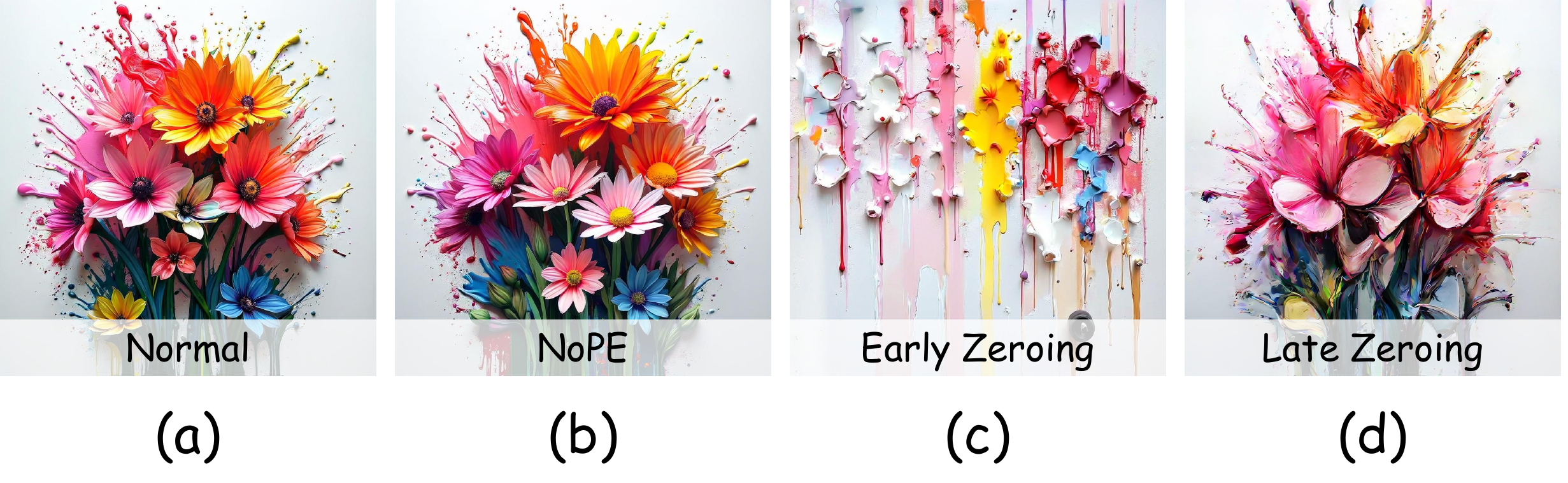}
        \caption{\small \textbf{Very low-frequency RoPE components exhibit near-NoPE behavior.}
        \emph{(a) Unperturbed image.}
        \emph{(b) NoPE substitution} on very low-frequency RoPE components preserves visual quality.
        \emph{(c) Q/K zeroing} during the Layout Construction Stage disrupts concept formation.
        \emph{(d) Q/K zeroing} during the Detail Refinement Stage blurs fine details.}
        \label{fig:abl_vlow}
    \end{minipage}%
    \hfill
    \begin{minipage}[t]{0.48\linewidth}
        \centering
        \includegraphics[width=\linewidth]{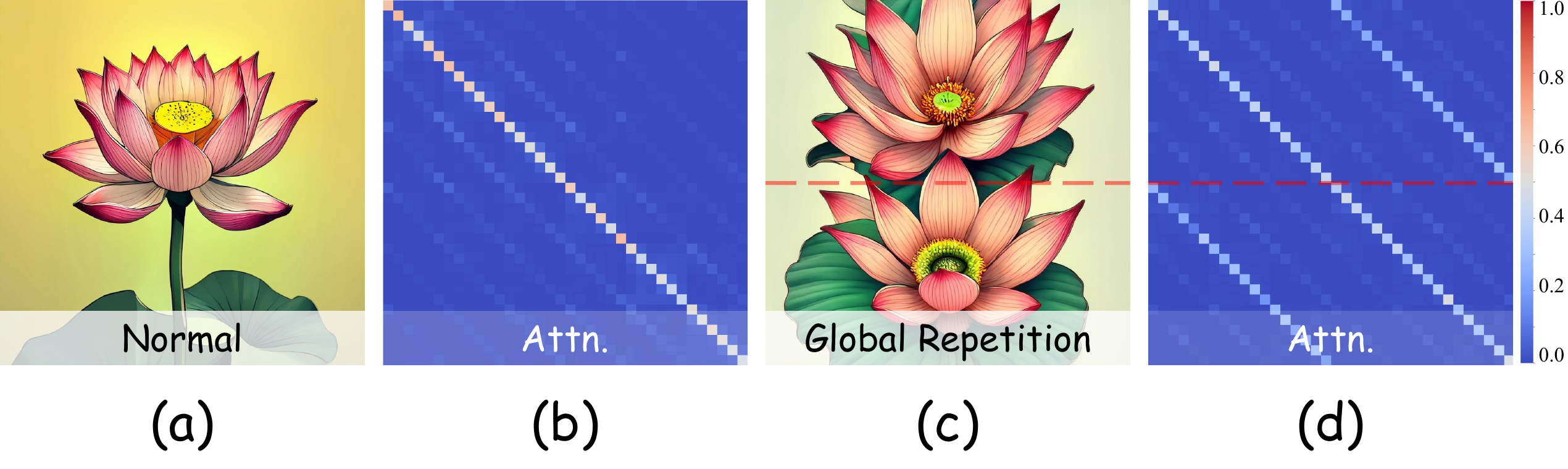}
        \caption{\small \textbf{Low-frequency components govern global layout in the Layout Construction Stage.}
        \emph{(a) Unperturbed image.}
        \emph{(b) Unperturbed attention map} with a typical diagonal pattern.
        \emph{(c) Early-stage NoPE} on these dimensions induces severe vertical global repetition.
        \emph{(d) Corresponding attention map} with near-identical upper and lower regions.}
        \label{fig:abl_low}
    \end{minipage}
\end{figure}

Directly scaling VAR training to much higher resolutions is generally impractical, as the sequence length grows rapidly and incurs prohibitive memory and optimization costs. Consequently, high-resolution synthesis typically relies on inference-time RoPE adaptation. Nevertheless, existing adaptation strategies exhibit clear limitations: PI~\citep{chen2023extending} tends to overly weaken high-frequency positional information, resulting in over-smoothed textures and loss of local details; NTK often introduces global repetition and semantic inconsistency, while YaRN~\citep{peng2023yarn}, although more robust in practice, still suffers from local repetition under large extrapolation ratios.

To investigate the underlying cause, we first partition the RoPE rotary dimension pairs by their wavelength $T_j$ relative to the side length $L$ of the training-time 2D token map into four bands: High (the $m$ pairs with smallest $T_j$), Mid ($T_j<L$), Low ($L\le T_j\le 4L$), and Very Low ($T_j>4L$). We then perform controlled dimension-wise RoPE interventions on these four frequency bands, either replacing selected dimensions with NoPE~\citep{kazemnejad2023impact} ($\theta_j \to 0$, $T_j \to \infty$) or modifying their wavelengths in different generation stages, and examine the resulting images and attention maps. The detailed analysis is as follows.

\smallskip
\noindent
\textit{\textbf{Observation 1 (Very low-frequency components behave similarly to NoPE).}}
As shown in Figure~\ref{fig:abl_vlow}\textcolor{cvprblue}{b}, replacing the very low-frequency RoPE dimensions with NoPE throughout generation still yields visually normal results, indicating that these dimensions exhibit near-NoPE behavior after training. To further probe their role, we additionally zero out the corresponding Q and K features at different stages. As shown in Figure~\ref{fig:abl_vlow}\textcolor{cvprblue}{c-d}, this intervention disrupts coherent concept formation when applied in the Layout Construction Stage, but induces noticeable blurring of fine details and edges when applied in the Detail Refinement Stage. These contrasting results indicate that, although very low-frequency components do not encode explicit positional information via rotation, they still provide essential auxiliary cues that jointly support global structure formation and fine detail rendering.

\smallskip
\noindent
\textit{\textbf{Observation 2 (Low-frequency components govern global structure in the Layout Construction Stage).}}
We replace low-frequency RoPE dimensions with NoPE in the Layout Construction Stage. As shown in Figure~\ref{fig:abl_low}\textcolor{cvprblue}{c}, the generated image exhibits clear global repetition along the vertical direction, and the corresponding attention map in Figure~\ref{fig:abl_low}\textcolor{cvprblue}{d} shows nearly identical patterns in the upper and lower regions. This indicates that low-frequency components encode the global layout; once they are removed, the model is forced to rely on higher-frequency components, whose shorter periods periodically tile the enlarged canvas with near-identical patterns, resulting in observed global repetition.

\begin{wrapfigure}{r}{0.5\linewidth}
    \centering
    \vspace{-12pt}
    \includegraphics[width=\linewidth]{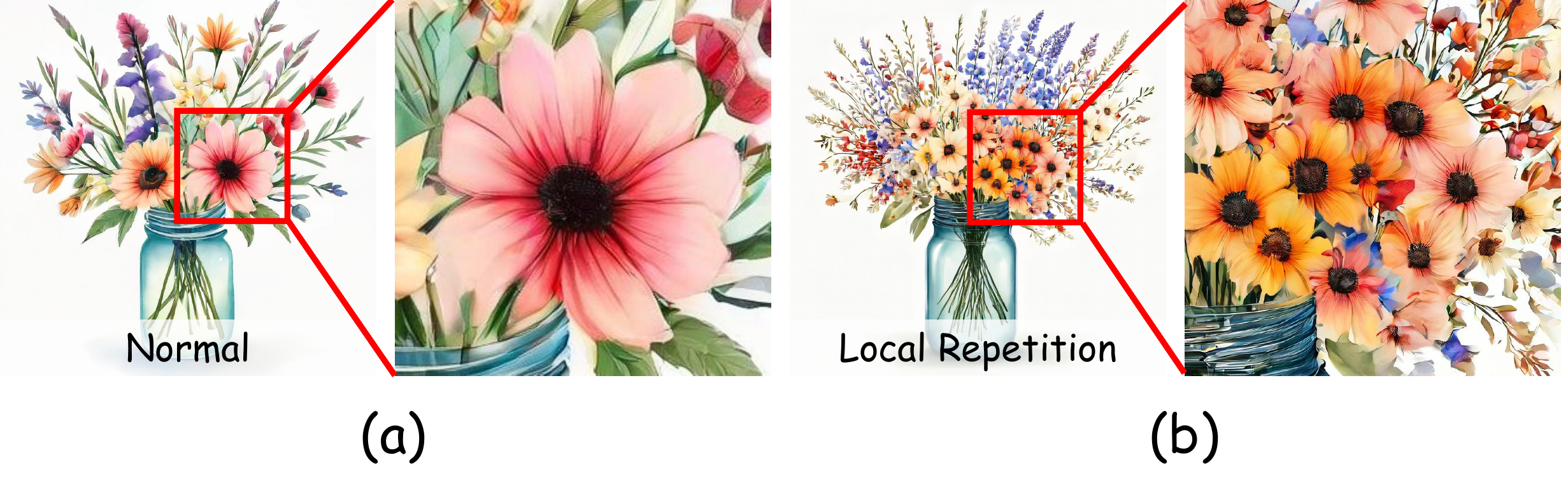}
    \vspace{-18pt}
    \caption{\small \textbf{Mid-frequency distortion causes local repetition in the Local Refinement Stage.}
    \emph{(a) Unperturbed image} with a detail crop at the typical element size.
    \emph{(b) Short-wavelength forcing} to $L/6$ induces local repetition with smaller and more numerous compositional elements.}
    \label{fig:abl_mid}
    \vspace{-12pt}
\end{wrapfigure}
\smallskip
\noindent
\textit{\textbf{Observation 3 (Mid-frequency components govern mid-sized object composition in the Local Refinement Stage).}}
We next study mid-frequency RoPE dimensions by forcing their wavelengths to a small value (\ie, $L/6$). As shown in Figure~\ref{fig:abl_mid}\textcolor{cvprblue}{b}, the generated image exhibits clear local repetition, with compositional elements appearing at noticeably smaller sizes and in greater numbers compared with the unperturbed image, as further confirmed by the zoomed-in crops. This indicates that mid-sized objects rely on mid-frequency components whose periods match their spatial scale; collapsing this band to an overly short wavelength misaligns the positional encoding with object scale, yielding local repetition and size inconsistency.

\begin{figure*}[h]
    \captionsetup{type=figure}%
    \includegraphics[width=\linewidth]{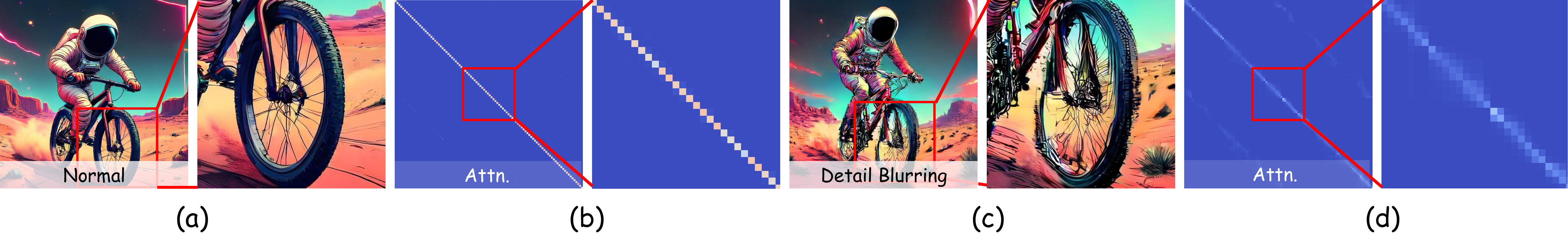}%
    \vspace{-0.4cm}
    \caption{\small \textbf{High-frequency components are critical for fine detail in the Detail Refinement Stage.}
    \emph{(a) Unperturbed image} with a detail crop showing sharp edges.
    \emph{(b) Corresponding attention map} with a typical diagonal pattern.
    \emph{(c) Late-stage NoPE} on these dimensions blurs edges and fine textures.
    \emph{(d) Corresponding attention map} with leakage around the diagonal.
    }%
    \label{fig:abl_high}%
    \label{fig:attn_high}%
    \vspace{-0.5cm}
\end{figure*}%

\smallskip
\noindent
\textit{\textbf{Observation 4 (High-frequency components are critical for fine detail generation).}}
Finally, we replace the high-frequency RoPE dimensions with NoPE in the Detail Refinement Stage. As shown in Figure~\ref{fig:abl_high}\textcolor{cvprblue}{c}, this causes clear detail degradation compared with the unperturbed image, with edges becoming blurred and fine textures lost. The attention map shown in Figure~\ref{fig:abl_high}\textcolor{cvprblue}{d} further supports this observation. We posit that the model relies on high-frequency dimensions to encode fine-grained neighboring positional relations in the Detail Refinement Stage; replacing them with NoPE directly undermines detail fidelity.

Taken together, the four observations reveal a clear band-stage mismatch: each stage is governed by a distinct dominant frequency band whose role further varies from stage to stage, so no static RoPE remapping can satisfy all three stages simultaneously. This stage-dependent behavior directly motivates the stage-aware design presented next.

\subsection{Stage-Aware RoPE Remapping}
\label{sec:dynamic_rope}

Since RoPE extrapolation in VAR is inherently stage-dependent, any fixed remapping is fundamentally unsuitable for progressive generation; we therefore propose a \emph{Stage-Aware RoPE Remapping} that adapts the positional frequency encoding to the distinct requirements of each generation stage.

We begin by introducing a general scale-step-dependent remapping formulation. Let $k$ denote the scale-step index; for the $j$-th feature pair, the frequency at scale step $k$ is defined as
\begin{equation}
\label{eq:scale_freq}
\theta_{k,j}=\phi_{k,j}(\theta_j).
\end{equation}
Accordingly, at token position $n$, the $j$-th feature pair is rotated by angle $n\theta_{k,j}$.

We then assign a tailored remapping to each band at each stage. Following Observation~1, very low-frequency dimensions are set to NoPE (\ie, $\phi_{k,j}(\theta_j)=0$) uniformly across all stages. For the remaining three bands: PI is applied in the Layout Construction Stage to scale frequencies proportionally with the target resolution, preserving the spatial proportions of layout elements; a linear interpolation between PI and YaRN is adopted in the Local Refinement Stage to transition from global-layout preservation toward more localized generation; and YaRN is used in the Detail Refinement Stage to retain fine-grained positional discrimination for high-frequency detail rendering. Formally, the unified remapping for the remaining three bands is defined as in Equation~\ref{eq:dynamic_rope_interp}:
\begin{equation}
\label{eq:dynamic_rope_interp}
\phi_{k,j}(\theta_j)
=
(1-\omega_k)\phi^{\mathrm{PI}}(\theta_j)
+
\omega_k \phi_j^{\mathrm{YaRN}}(\theta_j),
\end{equation}
where $\phi^{\mathrm{PI}}$ and $\phi_j^{\mathrm{YaRN}}$ denote the PI and YaRN frequency remappings, respectively (formally defined in Appendix~\ref{sec:rope_background}), and $\omega_k\in[0,1]$ is the stage-dependent interpolation weight: it is set to $0$ throughout the Layout Construction Stage, increases linearly from $0$ to $1$ across the Local Refinement Stage, and is set to $1$ throughout the Detail Refinement Stage. The thresholds $\lambda_{\mathrm{lo}}$ and $\lambda_{\mathrm{hi}}$ of YaRN are set to the High/Mid and Mid/Low band boundaries, respectively.

%%%%%%%%%%%%%%%%%%%%%%%%%%%%%%%%%%%%%%%%%%%%%%%%%%%%

\subsection{Entropy-Driven Adaptive Attention Calibration}
\label{sec:eaac}

As shown in Figure~\ref{fig:attn_high}, resolution extrapolation often leads to blurred fine details during the later stages of generation. Our Stage-Aware RoPE Remapping perturbs the queries and keys to mitigate this issue but does not explicitly control the concentration of the attention distribution. Some methods explicitly perturb attention: YaRN adopts a single global scaling factor to rescale attention, whereas UltraImage~\citep{zhao2025ultraimage} sharpens the attention distribution with head-specific scaling factors. However, both methods rely on predefined coefficients, neglecting the variability across different generation stages and target extrapolated resolutions.

To address this issue, we propose \emph{Entropy-Driven Adaptive Attention Calibration}. We are guided by the principle that \emph{the relative dispersion of attention distributions should remain approximately invariant across resolutions}; the main difficulty is to quantify dispersion in a manner that is comparable when the number of tokens $N$ changes. UltraImage measures dispersion with Shannon entropy, but raw entropy is not directly comparable across resolutions because its upper bound scales with $\log N$ as the number of tokens grows. We therefore adopt a normalized entropy formulation that admits direct cross-resolution comparison.

Specifically, let $Q^{\ell,h}\in\mathbb{R}^{N_q\times d}$ and $K^{\ell,h}\in\mathbb{R}^{N_k\times d}$ denote the query and key of head $h$ at layer $\ell$, with per-head dimension $d$.
For each attention head, standard multi-head attention first forms the pre-softmax attention logits $S^{\ell,h}$ and then obtains the corresponding attention map $P^{\ell,h}$ via row-wise softmax:
\begin{equation}
\label{eq:mha_recap}
S^{\ell,h} = \frac{Q^{\ell,h}(K^{\ell,h})^{\top}}{\sqrt{d}},\qquad
P^{\ell,h} = \mathrm{Softmax}(S^{\ell,h}),
\end{equation}
where $S^{\ell,h},\,P^{\ell,h}\in\mathbb{R}^{N_q\times N_k}$. For notational simplicity, we omit the superscript $(\ell,h)$ in the following section. We assign each head a dedicated scaling factor and compute the modulated attention as:
\begin{equation}
\label{eq:dsattn}
P(\alpha)=\mathrm{Softmax}(\alpha S),
\end{equation}
where $\alpha$ denotes the scaling factor associated with layer $\ell$ and head $h$. We then define the normalized attention entropy as:
\begin{equation}
\label{eq:norm-entropy}
\mathcal{H}(\alpha)= -\frac{1}{N_q\log N_k}\sum_{i=1}^{N_q}\sum_{j=1}^{N_k} P_{ij}(\alpha)\log P_{ij}(\alpha).
\end{equation}
It is straightforward to show that $\mathcal{H}(\alpha)\in[0,1]$, where smaller values indicate more concentrated attention distributions.

Let $\mathcal{H}^{\mathrm{tr}}$ denote the entropy at the training resolution when $\alpha=1$. Unless otherwise specified, $\mathcal{H}(\alpha)$ denotes the normalized attention entropy evaluated at the extrapolated resolution in the following section. Then the gap $\mathcal{H}(1)-\mathcal{H}^{\mathrm{tr}}$ measures how far the current attention map departs from its normal concentration level. Our calibration restores this concentration by adjusting the per-head scaling factor $\alpha$ so that $\mathcal{H}(\alpha)$ matches its training-resolution counterpart $\mathcal{H}^{\mathrm{tr}}$:
\begin{equation}
\label{eq:target}
\mathcal{H}(\alpha)=\mathcal{H}^{\mathrm{tr}}.
\end{equation}

Solving Equation~\ref{eq:target} exactly for $\alpha$ is computationally expensive. To obtain an efficient approximation, we linearize $\mathcal{H}(\alpha)$ around $\alpha=1$. Let $P_i(\alpha)$ and $S_i$ denote the $i$-th rows of $P(\alpha)$ and $S$, respectively. To characterize how the entropy varies with the scaling factor, we differentiate $\mathcal{H}(\alpha)$ with respect to $\alpha$ and obtain:
\begin{equation}
\label{eq:dHcal}
\frac{d\mathcal{H}(\alpha)}{d\alpha}
=
-\frac{\alpha}{N_q\log N_k}\sum_{i=1}^{N_q}\mathrm{Var}_{P_i(\alpha)}(S_i)
\le 0.
\end{equation}
This shows that increasing $\alpha$ sharpens the attention distribution. For convenience, we define:
\begin{equation}
\label{eq:Vg-def}
V_g=\frac{1}{N_q}\sum_{i=1}^{N_q}\mathrm{Var}_{P_i(1)}(S_i).
\end{equation}
We then apply the first-order Taylor approximation of $\mathcal{H}(\alpha)$ at $\alpha=1$ to obtain:
\begin{equation}
\mathcal{H}(\alpha)\approx \mathcal{H}(1)-\frac{V_g}{\log N_k}(\alpha-1).
\end{equation}
By substituting this approximation into Equation~\ref{eq:target}, we obtain the closed-form estimate of the scaling factor in Equation~\ref{eq:alpha-init}:
\begin{equation}
\label{eq:alpha-init}
\hat{\alpha}
=
1+\frac{\bigl(\mathcal{H}(1)-\mathcal{H}^{\mathrm{tr}}\bigr)\log N_k}{V_g}.
\end{equation}
We use $\hat{\alpha}$ as the scaling factor for each selected head. In practice, we apply this calibration only when $k>k_h$ and $\mathcal{H}^{\mathrm{tr}}<\tau_h$; otherwise, we set $\alpha=1$. To avoid numerical instability when $V_g$ is close to zero, we additionally fall back to $\alpha=1$ whenever $V_g$ is below a small threshold $\epsilon$.
\section{Experiments}
\label{sec:experiments}

\subsection{Experimental Setup}

\textbf{Models and Evaluations.}
We evaluate our method on Infinity~\citep{han2025infinity}, a VAR-based text-to-image model whose native maximum resolution is $1024 \times 1024$. At inference time, we extrapolate the model to $2560 \times 2560$ for quantitative evaluation, which is the maximum resolution allowed by our hardware budget, and additionally produce qualitative results at $2048 \times 2048$. We compare against representative positional extrapolation methods, including PE, PI~\citep{chen2023extending}, NTK-aware scaling, YaRN~\citep{peng2023yarn}, RiFlex~\citep{zhao2025riflex}, and DyPE~\citep{issachar2025dype}. To reduce memory usage and push evaluation to the highest feasible resolution, we perform inference with FastVAR~\citep{guo2025fastvar}. We evaluate generation quality using GenEval~\citep{ghosh2023geneval}, DPG-Bench~\citep{hu2024ella}, and HPSv2.1~\citep{wu2023human}.

\textbf{Implementation Details.}
For the Stage-Aware RoPE Remapping, we use a total of $K=13$ generation scale steps and set $k_l=6$ and $k_h=9$ for stage-aware interpolation between PI and YaRN, and the size of the High band is set to $m=3$. For the Attention Calibration, the reference entropy is measured at the training resolution and reused during extrapolated inference. We activate Attention Calibration only when $k>k_h$ and $\mathcal{H}^{\mathrm{tr}}<\tau_h$ with $\tau_h=0.3$. Unless otherwise specified, all remaining hyperparameter settings follow the default configuration of Infinity. All experiments are conducted on a single NVIDIA RTX A6000 Ada GPU.

\subsection{Qualitative Analysis}
\begin{figure*}[h]
    \flushleft
    \includegraphics[width=\linewidth]{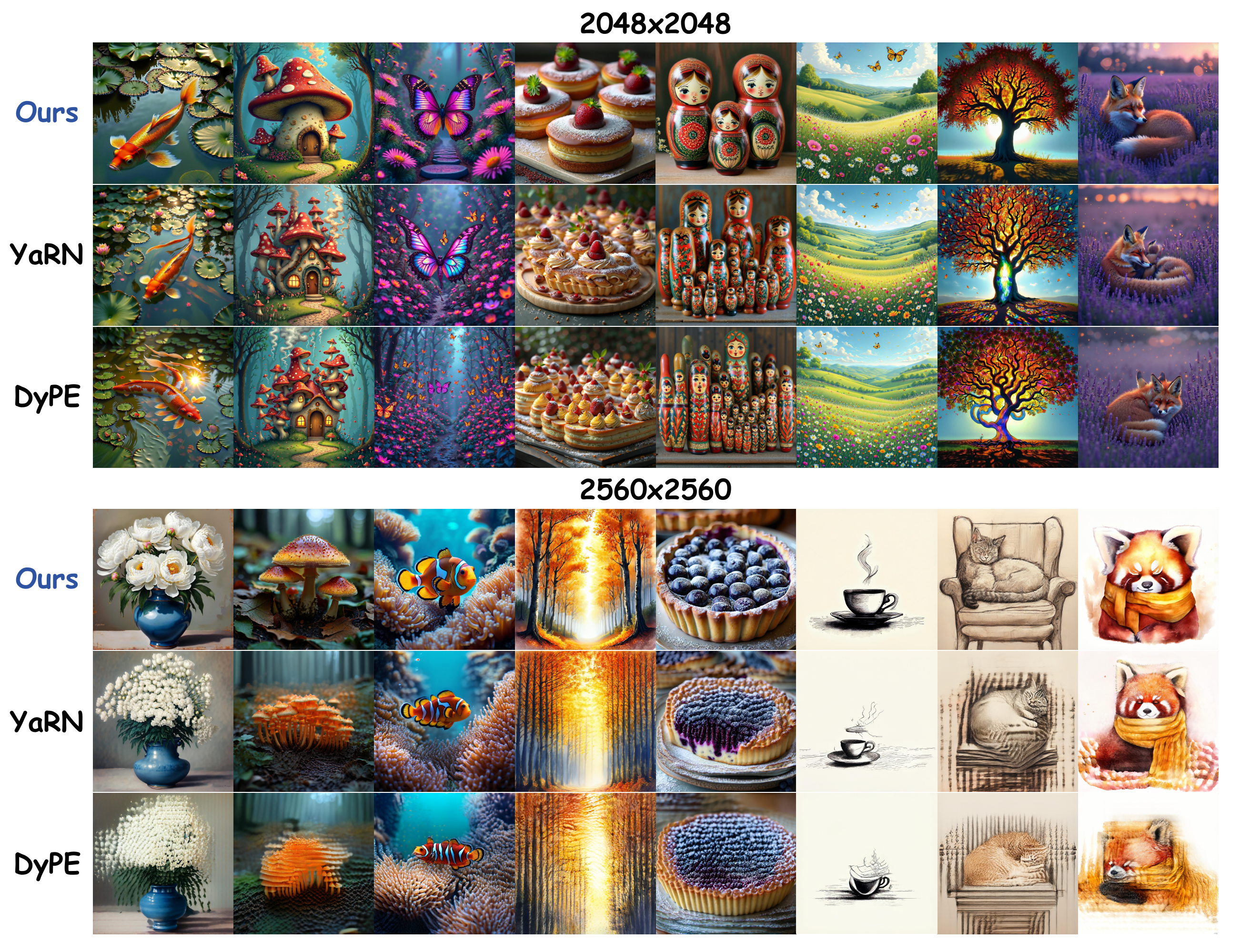}
    \vspace{-0.7cm}
    \caption{\small \textbf{Qualitative comparison of ExtraVAR against representative resolution-extrapolation baselines.}}
    \label{fig:visualization}
\end{figure*}

As shown in Figure~\ref{fig:visualization}, we compare our method with YaRN and DyPE under identical experimental settings at $2048\times2048$ and $2560\times2560$ resolutions. Despite generating images with up to $\sim 6.25\times$ more pixels than the original training resolution, our method consistently preserves high visual quality and strong semantic fidelity. In particular, it maintains coherent global structure, clear object boundaries, and plausible fine-grained details, without introducing obvious artifacts during resolution extrapolation. By contrast, the images produced by YaRN and DyPE exhibit noticeable degradation, including local repetition, structural distortion, and loss of detail, especially in spatially complex regions. These qualitative results indicate that our method is substantially more effective at handling extreme resolution extrapolation, leading to more faithful and visually realistic generations under the same conditions.

\begin{table*}[t]
\centering
\caption{\textbf{Quantitative comparison on GenEval~\citep{ghosh2023geneval} and DPG-Bench~\citep{hu2024ella}.} GenEval columns: S.Obj.\ (single object), T.Obj.\ (two objects), Cnt.\ (counting), Col.\ (colors), Pos.\ (position), C.Atr.\ (color attribution), Ov.\ (overall). DPG-Bench columns: Glo.\ (global), Ent.\ (entity), Rel.\ (relation), Atr.\ (attribute), Ov.\ (overall). ``--'' marks missing entries.}
\label{tab:gen_model_comparison}
\footnotesize
\setlength{\tabcolsep}{1.6pt}
\setlength{\extrarowheight}{0.5pt}
\resizebox{\linewidth}{!}{%
\begin{tabular}{lccccccc|ccccc}
\toprule
\multirow{2}{*}{\textbf{Model}} & \multicolumn{7}{c|}{\textbf{GenEval}} & \multicolumn{5}{c}{\textbf{DPG-Bench}} \\
\cmidrule(lr){2-8} \cmidrule(lr){9-13}
 & \textbf{S.Obj.}$\uparrow$ & \textbf{T.Obj.}$\uparrow$ & \textbf{Cnt.}$\uparrow$ & \textbf{Col.}$\uparrow$ & \textbf{Pos.}$\uparrow$ & \textbf{C.Atr.}$\uparrow$ & \textbf{Ov.}$\uparrow$ & \textbf{Glo.}$\uparrow$ & \textbf{Ent.}$\uparrow$ & \textbf{Rel.}$\uparrow$ & \textbf{Atr.}$\uparrow$ & \textbf{Ov.}$\uparrow$ \\
\midrule
SDXL~\citep{podell2023sdxl} & 0.98 & 0.74 & 0.39 & 0.85 & 0.15 & 0.23 & 0.55 & 83.27 & 82.43 & 86.76 & 80.91 & 74.65 \\
LlamaGen~\citep{sun2024autoregressive} & 0.71 & 0.34 & 0.21 & 0.58 & 0.07 & 0.04 & 0.32 & -- & -- & -- & -- & 65.16 \\
Show-o~\citep{xie2024show} & 0.98 & 0.80 & 0.66 & 0.84 & 0.31 & 0.50 & 0.68 & 79.33 & 75.44 & 84.45 & 78.02 & 67.27 \\
PixArt-Sigma~\citep{chen2024pixart} & 0.98 & 0.59 & 0.50 & 0.80 & 0.10 & 0.15 & 0.52 & 86.89 & 82.89 & 86.59 & 88.94 & 80.54 \\
HART~\citep{tang2024hart} & -- & -- & -- & -- & -- & -- & 0.56 & -- & -- & -- & -- & 80.89 \\
DALL-E 3 & 0.96 & 0.87 & 0.47 & 0.83 & 0.43 & 0.45 & 0.67 & 90.97 & 89.61 & 90.58 & 88.39 & 83.50 \\
Emu3~\citep{wang2024emu3} & 0.99 & 0.81 & 0.42 & 0.80 & 0.49 & 0.45 & 0.66 & -- & 87.17 & 90.61 & 86.33 & 81.60 \\
\midrule
\textcolor{Grey}{PI~\citep{chen2023extending}} & \textcolor{Grey}{0.95} & \textcolor{Grey}{0.62} & \textcolor{Grey}{0.45} & \textcolor{Grey}{0.73} & \textcolor{Grey}{0.27} & \textcolor{Grey}{0.40} & \textcolor{Grey}{0.57} & \textcolor{Grey}{81.46} & \textcolor{Grey}{89.13} & \textcolor{Grey}{93.14} & \textcolor{Grey}{85.52} & \textcolor{Grey}{82.69} \\
\graymidrule
PE & $0.10_{\textcolor{DarkGrey}{-0.85}}$ & $0.00_{\textcolor{DarkGrey}{-0.62}}$ & $0.00_{\textcolor{DarkGrey}{-0.45}}$ & $0.03_{\textcolor{DarkGrey}{-0.70}}$ & $0.00_{\textcolor{DarkGrey}{-0.27}}$ & $0.00_{\textcolor{DarkGrey}{-0.40}}$ & $0.02_{\textcolor{DarkGrey}{-0.55}}$ & $68.09_{\textcolor{DarkGrey}{-13.37}}$ & $37.19_{\textcolor{DarkGrey}{-51.94}}$ & $72.54_{\textcolor{DarkGrey}{-20.60}}$ & $63.25_{\textcolor{DarkGrey}{-22.27}}$ & $24.88_{\textcolor{DarkGrey}{-57.81}}$ \\
YaRN~\citep{peng2023yarn} & $0.99_{\textcolor{LightGreen}{+0.04}}$ & $0.83_{\textcolor{LightGreen}{+0.21}}$ & $0.28_{\textcolor{DarkGrey}{-0.17}}$ & $0.84_{\textcolor{LightGreen}{+0.11}}$ & $0.37_{\textcolor{LightGreen}{+0.10}}$ & $0.58_{\textcolor{LightGreen}{+0.18}}$ & $0.65_{\textcolor{LightGreen}{+0.08}}$ & $81.76_{\textcolor{LightGreen}{+0.30}}$ & $86.31_{\textcolor{DarkGrey}{-2.82}}$ & $92.63_{\textcolor{DarkGrey}{-0.51}}$ & $84.92_{\textcolor{DarkGrey}{-0.60}}$ & $78.70_{\textcolor{DarkGrey}{-3.99}}$ \\
RiFlex~\citep{zhao2025riflex} & $0.82_{\textcolor{DarkGrey}{-0.13}}$ & $0.42_{\textcolor{DarkGrey}{-0.20}}$ & $0.11_{\textcolor{DarkGrey}{-0.34}}$ & $0.61_{\textcolor{DarkGrey}{-0.12}}$ & $0.19_{\textcolor{DarkGrey}{-0.08}}$ & $0.18_{\textcolor{DarkGrey}{-0.22}}$ & $0.39_{\textcolor{DarkGrey}{-0.18}}$ & $75.38_{\textcolor{DarkGrey}{-6.08}}$ & $71.21_{\textcolor{DarkGrey}{-17.92}}$ & $87.54_{\textcolor{DarkGrey}{-5.60}}$ & $77.73_{\textcolor{DarkGrey}{-7.79}}$ & $60.48_{\textcolor{DarkGrey}{-22.21}}$ \\
DyPE~\citep{issachar2025dype} & $0.91_{\textcolor{DarkGrey}{-0.04}}$ & $0.60_{\textcolor{DarkGrey}{-0.02}}$ & $0.13_{\textcolor{DarkGrey}{-0.32}}$ & $0.76_{\textcolor{LightGreen}{+0.03}}$ & $0.20_{\textcolor{DarkGrey}{-0.07}}$ & $0.31_{\textcolor{DarkGrey}{-0.09}}$ & $0.49_{\textcolor{DarkGrey}{-0.08}}$ & $82.67_{\textcolor{LightGreen}{+1.21}}$ & $77.52_{\textcolor{DarkGrey}{-11.61}}$ & $89.16_{\textcolor{DarkGrey}{-3.98}}$ & $81.49_{\textcolor{DarkGrey}{-4.03}}$ & $68.65_{\textcolor{DarkGrey}{-14.04}}$ \\
\rowcolor{gray!15}
\textbf{Ours} & $0.99_{\textcolor{LightGreen}{+0.04}}$ & $0.83_{\textcolor{LightGreen}{+0.21}}$ & $0.57_{\textcolor{LightGreen}{+0.12}}$ & $0.84_{\textcolor{LightGreen}{+0.11}}$ & $0.38_{\textcolor{LightGreen}{+0.11}}$ & $0.59_{\textcolor{LightGreen}{+0.19}}$ & $0.70_{\textcolor{LightGreen}{+0.13}}$ & $83.28_{\textcolor{LightGreen}{+1.82}}$ & $89.50_{\textcolor{LightGreen}{+0.37}}$ & $93.02_{\textcolor{DarkGrey}{-0.12}}$ & $86.83_{\textcolor{LightGreen}{+1.31}}$ & $83.58_{\textcolor{LightGreen}{+0.89}}$ \\
\bottomrule
\end{tabular}%
}
\end{table*}

\subsection{Quantitative Evaluations}
\begin{figure}[t]
    \centering
    \begin{minipage}[t]{0.42\linewidth}
        \centering
        \vspace{0pt}
        \includegraphics[width=\linewidth]{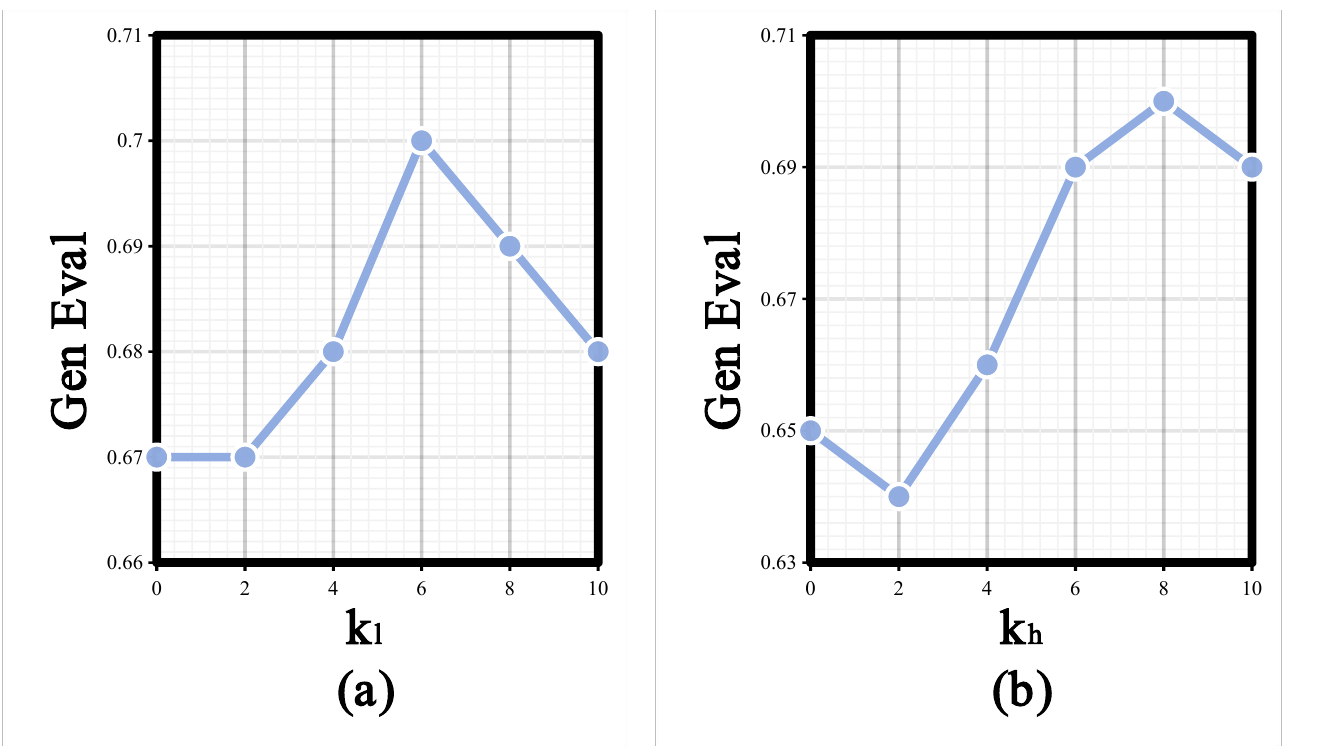}
        \caption{\textbf{Sensitivity analysis of hyperparameters $k_l$ and $k_h$.}}
        \label{fig:scaling}
    \end{minipage}%
    \hfill
    \begin{minipage}[t]{0.55\linewidth}
        \centering
        \vspace{0pt}
        \captionof{table}{\textbf{Comparison on HPSv2.1~\citep{wu2023human}.} Style columns: Photo (photo), C-Art (concept art), Anime (anime), Paint.\ (paintings); Ov.\ denotes the overall score.}
        \label{tab:hps_model_comparison}
        \footnotesize
        \setlength{\tabcolsep}{2.2pt}
        \setlength{\extrarowheight}{0.5pt}
        \resizebox{\linewidth}{!}{%
        \begin{tabular}{lccccc}
        \toprule
        \textbf{Model} & \textbf{Photo}$\uparrow$ & \textbf{C-Art}$\uparrow$ & \textbf{Anime}$\uparrow$ & \textbf{Paint.}$\uparrow$ & \textbf{Ov.}$\uparrow$ \\
        \midrule
        \textcolor{Grey}{PI~\citep{chen2023extending}} & \textcolor{Grey}{26.80} & \textcolor{Grey}{28.19} & \textcolor{Grey}{29.40} & \textcolor{Grey}{27.60} & \textcolor{Grey}{28.00} \\
        \graymidrule
        PE & $9.14_{\textcolor{DarkGrey}{-17.66}}$ & $8.75_{\textcolor{DarkGrey}{-19.44}}$ & $8.55_{\textcolor{DarkGrey}{-20.85}}$ & $9.20_{\textcolor{DarkGrey}{-18.40}}$ & $8.91_{\textcolor{DarkGrey}{-19.09}}$ \\
        YaRN~\citep{peng2023yarn} & $27.91_{\textcolor{LightGreen}{+1.11}}$ & $28.20_{\textcolor{LightGreen}{+0.01}}$ & $30.04_{\textcolor{LightGreen}{+0.64}}$ & $27.43_{\textcolor{DarkGrey}{-0.17}}$ & $28.40_{\textcolor{LightGreen}{+0.40}}$ \\
        RiFlex~\citep{zhao2025riflex} & $16.51_{\textcolor{DarkGrey}{-10.29}}$ & $14.96_{\textcolor{DarkGrey}{-13.23}}$ & $15.57_{\textcolor{DarkGrey}{-13.83}}$ & $14.35_{\textcolor{DarkGrey}{-13.25}}$ & $15.35_{\textcolor{DarkGrey}{-12.65}}$ \\
        DyPE~\citep{issachar2025dype} & $20.12_{\textcolor{DarkGrey}{-6.68}}$ & $17.82_{\textcolor{DarkGrey}{-10.37}}$ & $20.67_{\textcolor{DarkGrey}{-8.73}}$ & $17.41_{\textcolor{DarkGrey}{-10.19}}$ & $19.00_{\textcolor{DarkGrey}{-9.00}}$ \\
        \rowcolor{gray!15}
        \textbf{Ours} & $29.24_{\textcolor{LightGreen}{+2.44}}$ & $30.04_{\textcolor{LightGreen}{+1.85}}$ & $31.51_{\textcolor{LightGreen}{+2.11}}$ & $29.64_{\textcolor{LightGreen}{+2.04}}$ & $30.11_{\textcolor{LightGreen}{+2.11}}$ \\
        \bottomrule
        \end{tabular}%
        }
    \end{minipage}
\end{figure}

\begin{table*}[t]
  \caption{\textbf{Ablation of the proposed Stage-Aware RoPE Remapping and Entropy-Driven Attention Calibration on GenEval~\citep{ghosh2023geneval}.}}
  \label{tab:ablation}
  \centering
  \footnotesize
  \setlength{\tabcolsep}{3pt}
  \resizebox{\linewidth}{!}{%
  \begin{tabular}{cc|ccccccc}
  \toprule
  \textbf{Stage-Aware RoPE} & \textbf{Attention Calibration} & \textbf{Single Obj.}$\uparrow$ & \textbf{Two Obj.}$\uparrow$ & \textbf{Counting}$\uparrow$ & \textbf{Colors}$\uparrow$ & \textbf{Position}$\uparrow$ & \textbf{Color Attri.}$\uparrow$ & \textbf{Overall}$\uparrow$ \\
  \midrule
  \xmark & \xmark & 0.95 & 0.62 & 0.45 & 0.73 & 0.27 & 0.40 & 0.57 \\
  \midrule
  \xmark & \checkmark & $0.94_{\textcolor{DarkGrey}{-0.01}}$ & $0.62_{\textcolor{DarkGrey}{+0.00}}$ & $0.45_{\textcolor{DarkGrey}{+0.00}}$ & $0.77_{\textcolor{LightGreen}{+0.04}}$ & $0.27_{\textcolor{DarkGrey}{+0.00}}$ & $0.42_{\textcolor{LightGreen}{+0.02}}$ & $0.58_{\textcolor{LightGreen}{+0.01}}$ \\
  \checkmark & \xmark & $0.99_{\textcolor{LightGreen}{+0.04}}$ & $0.82_{\textcolor{LightGreen}{+0.20}}$ & $0.58_{\textcolor{LightGreen}{+0.13}}$ & $0.85_{\textcolor{LightGreen}{+0.12}}$ & $0.35_{\textcolor{LightGreen}{+0.08}}$ & $0.53_{\textcolor{LightGreen}{+0.13}}$ & $0.69_{\textcolor{LightGreen}{+0.12}}$ \\
  \checkmark & \checkmark & $0.99_{\textcolor{LightGreen}{+0.04}}$ & $0.83_{\textcolor{LightGreen}{+0.21}}$ & $0.57_{\textcolor{LightGreen}{+0.12}}$ & $0.84_{\textcolor{LightGreen}{+0.11}}$ & $0.38_{\textcolor{LightGreen}{+0.11}}$ & $0.59_{\textcolor{LightGreen}{+0.19}}$ & $0.70_{\textcolor{LightGreen}{+0.13}}$ \\
  \bottomrule
  \end{tabular}%
  }
\end{table*}

\textbf{Main results on GenEval and DPG-Bench.}
Table~\ref{tab:gen_model_comparison} summarizes the quantitative results under our resolution-extrapolation setting. ExtraVAR attains the best overall performance on both GenEval and DPG-Bench, indicating that the proposed method strengthens both compositional alignment and dense prompt following well beyond the native training resolution. On GenEval, it improves over PI~\citep{chen2023extending} by 0.13 in the overall score, with especially clear gains on Two Object, Counting, Position, and Color Attribution while remaining competitive on the remaining sub-tasks, suggesting that the improvement is broad rather than confined to a narrow subset. A similar pattern emerges on DPG-Bench~\citep{hu2024ella}, where ExtraVAR improves the overall score by 0.89 over PI~\citep{chen2023extending}, with consistent gains on the global, entity, and attribute dimensions and a nearly unchanged relation score. As DPG-Bench emphasizes fine-grained prompt fidelity under complex descriptions, these gains suggest that ExtraVAR not only suppresses visual artifacts but also preserves semantic faithfulness during high-resolution autoregressive generation.

\textbf{Quantitative Comparison on HPSv2.}
Table~\ref{tab:hps_model_comparison} further compares different methods on HPSv2.1~\citep{wu2023human}. ExtraVAR achieves the best overall score and improves over PI~\citep{chen2023extending} by 2.11, with consistent gains across all four style categories. This result suggests that the proposed method improves not only prompt alignment but also perceptual quality and human-preferred visual appearance under resolution extrapolation. The improvement across both photographic and artistic styles further indicates that the benefit is not restricted to a specific image domain, but reflects a more general improvement in image quality. By contrast, several extrapolation baselines show clear degradation on HPSv2.1~\citep{wu2023human}. For example, PE and RiFlex trail PI by 19.09 and 12.65 in the overall score, respectively.

\subsection{Ablation Study}

\paragraph{Impact of the stage transition indices $k_l$ and $k_h$.}
Figure~\ref{fig:scaling} examines the GenEval score against $k_l$ and $k_h$. The score is maximized when $k_l$ falls near the Layout Construction/Local Refinement boundary and $k_h$ near the Local Refinement/Detail Refinement boundary, and degrades on either side of these stage boundaries. This confirms that the three-stage partition of VAR generation is intrinsic to our Stage-Aware RoPE Remapping.

\textbf{Effectiveness of the Stage-Aware RoPE Remapping and Attention Calibration.}
Table~\ref{tab:ablation} validates the effectiveness of both proposed components. The Stage-Aware RoPE Remapping yields consistent gains across nearly all sub-tasks under resolution extrapolation, and incorporating the Attention Calibration further strengthens Position and Color Attribution while attaining the best overall score. The two components are therefore complementary, jointly enhancing positional adaptation and attention quality during inference.

\section{Conclusion}
\label{sec:conclusion}

In this paper, we introduced \textbf{ExtraVAR}, a training-free framework for resolution extrapolation in Visual Autoregressive (VAR) models. Through controlled dimension-wise RoPE interventions, we traced three characteristic failure modes of VAR extrapolation, namely global repetition, local repetition, and detail degradation, to a unified \emph{band-stage mismatch} in which each generation stage is governed by a distinct dominant RoPE frequency band. Building on this insight, we proposed \textbf{Stage-Aware RoPE Remapping}, which assigns each frequency band a stage-specific remapping rule that keeps the active remapping aligned with the dominant band of every stage. We further developed \textbf{Entropy-Driven Adaptive Attention Calibration}, which leverages a resolution-invariant normalized entropy and yields a closed-form per-head scaling factor that realigns the extrapolated-resolution attention with its training-resolution counterpart. Comprehensive experiments on GenEval, DPG-Bench, and HPSv2.1 show that ExtraVAR consistently surpasses prior baselines under extreme resolution extrapolation, advancing high-fidelity VAR generation beyond the native training scale.

%%%%%%%%%%%%%%%%%%%%%%%%%%%%%% Bibliography %%%%%%%%%%%%%%%%%%%%%%%%%%%%%%%%%%%%%
% Force the references to start on a fresh page so they never share a page with
% the conclusion, which makes the 9-page main-paper boundary unambiguous for reviewers.
\clearpage
\small
\bibliographystyle{ieeenat_fullname}
\bibliography{main}
\normalsize

%%%%%%%%%%%%%%%%%%%%%%%%%%%% Technical appendix %%%%%%%%%%%%%%%%%%%%%%%%%%%%%%%%%
\appendix
\section{Background}

\subsection{Visual autoregressive modeling via next-scale prediction}
\label{sec:var_background}

Visual autoregressive modeling (VAR) reformulates image autoregression from token-wise next-token prediction to scale-wise next-scale prediction. Let $(r_1,\ldots,r_K)$ denote multi-scale token maps ordered from coarse to fine, where $r_k\in\mathcal{V}^{h_k\times w_k}$. The generative process is factorized as:
\begin{equation}
\label{eq:var_factorization}
p(r_1,\ldots,r_K)=\prod_{k=1}^{K}p\bigl(r_k\mid r_{<k}\bigr),
\end{equation}
where $r_{<k}=(r_1,\ldots,r_{k-1})$ denotes the token maps at all preceding scale steps. At scale step $k$, VAR predicts the full token map $r_k$ in parallel conditioned on $r_{<k}$, resulting in a coarse-to-fine generation process in which earlier scale steps establish the global structure and later scale steps progressively refine local details. During training, block-causal masking enforces conditioning only on valid previous scale steps, while inference proceeds sequentially across scale steps with KV caching.

\subsection{Attention mechanism}
\label{sec:attention_background}

\emph{Multi-head attention} is a core component of transformer architectures for modeling token dependencies. It computes attention through multiple parallel heads, allowing the model to capture different interaction patterns across tokens. Let $Q^{\ell,h}\in\mathbb{R}^{N_q\times d}$, $K^{\ell,h}\in\mathbb{R}^{N_k\times d}$, and $V^{\ell,h}\in\mathbb{R}^{N_k\times d}$ denote the query, key, and value of head $h$ at layer $\ell$. The attention map for each head is:
\begin{equation}
\label{eq:attention_background}
S^{\ell,h}=\frac{Q^{\ell,h}(K^{\ell,h})^\top}{\sqrt{d}},\qquad
P^{\ell,h}=\mathrm{Softmax}(S^{\ell,h}),
\end{equation}
where $P^{\ell,h}\in\mathbb{R}^{N_q\times N_k}$. The output of this head is:
\begin{equation}
\label{eq:attention_output_background}
O^{\ell,h}=P^{\ell,h}V^{\ell,h}.
\end{equation}
The outputs of all heads are then concatenated and projected to produce the layer output:
\begin{equation}
\label{eq:mha_background}
O^{\ell}
=
\mathrm{Concat}\!\left(O^{\ell,1},\ldots,O^{\ell,H}\right)W_O^{\ell}
\end{equation}
where $W_O^{\ell}\in\mathbb{R}^{Hd\times d_{\mathrm{model}}}$ is the output projection matrix, and $d_{\mathrm{model}}$ denotes the hidden dimension of the transformer layer.

\subsection{Rotary position embedding (RoPE) and extrapolation}
\label{sec:rope_background}

\emph{Rotary position embedding (RoPE)}~\citep{su2024roformer} encodes positional information by rotating each 2D feature pair with a frequency-dependent phase. Let $d$ denote the per-head feature dimension. For a 1D position index $n$, the $j$-th feature pair is rotated by angle $n\theta_j$, where the angular frequency is given by:
\begin{equation}
\label{eq:rope_theta_background}
\theta_j=b^{-2(j-1)/d},\qquad j=1,\ldots,d/2,
\end{equation}
where $b$ is the RoPE base (we follow the default $b=10000$ used in Infinity). The corresponding wavelength is:
\begin{equation}
\label{eq:rope_period_background}
T_j=\frac{2\pi}{\theta_j}.
\end{equation}
In vision transformers, a common 2D extension applies RoPE separately along the height and width axes, rotating the corresponding query and key components before attention computation.

\emph{No Positional Encodings (NoPE)}~\citep{kazemnejad2023impact} denotes a setting where self-attention is used without explicit positional encoding. In our notation, it can be regarded as the degenerate case of RoPE with $\theta_j = 0$ for all rotary pairs, equivalently $T_j \to \infty$. Positional information is therefore learned implicitly through the attention structure.

\emph{Position Interpolation (PI)}~\citep{chen2023extending} rescales positions by a constant factor. For extrapolation from training length $L$ to a target length $L'$, we define the extrapolation ratio $s=L'/L$. PI is equivalently reflected as a uniform scaling of all RoPE frequencies:
\begin{equation}
\label{eq:pi_background}
\phi^{\mathrm{PI}}(\theta_j)=\frac{\theta_j}{s}.
\end{equation}
This preserves coarse geometric structure under extrapolation, but under large scaling factors it can over-compress high-frequency components.

\emph{NTK-aware scaling} instead performs a non-uniform modification by rescaling the RoPE base. One common form is:
\begin{equation}
\label{eq:ntk_background}
\phi^{\mathrm{NTK}}(\theta_j)
=
\lambda_{\mathrm{NTK}}^{-2(j-1)/d}\,\theta_j,
\qquad
\lambda_{\mathrm{NTK}}=s^{d/(d-2)}.
\end{equation}
Compared with PI, this adjustment preserves relatively more of the original high-frequency content while progressively approaching PI toward lower-frequency components.

\emph{YaRN}~\citep{peng2023yarn} further refines this idea through a frequency-dependent interpolation strategy. Unlike PI, YaRN does not apply a single shared remapping to all frequency pairs. Instead, it assigns each pair a dimension-dependent mixing coefficient $\rho_j^{\mathrm{YaRN}}$, yielding the following frequency remapping:
\begin{equation}
\label{eq:yarn_theta_short}
\phi_j^{\mathrm{YaRN}}(\theta_j)
=
\rho_j^{\mathrm{YaRN}}\,\theta_j
+
\bigl(1-\rho_j^{\mathrm{YaRN}}\bigr)\frac{\theta_j}{s},
\end{equation}
where $\rho_j^{\mathrm{YaRN}}$ is a dimension-dependent interpolation coefficient. Thus, YaRN behaves similarly to the original RoPE on high-frequency components while approaching PI toward lower-frequency components. YaRN also introduces a shared attention rescaling, which can be written as:
\begin{equation}
\label{eq:yarn_alpha_short}
P^{\ell,h}(\alpha_{\mathrm{global}})
=
\mathrm{Softmax}\!\left(\alpha_{\mathrm{global}} S^{\ell,h}\right),
\end{equation}
where $\alpha_{\mathrm{global}}$ is a shared hyperparameter across layers and heads.

\end{document}